\documentclass{article}
\usepackage{amsmath}
\usepackage{amssymb}
\usepackage{dsfont}
\usepackage{graphicx}
\usepackage{xcolor}

\title{Reverse Annealing for Nonnegative/Binary Matrix Factorization}
\author{John K. Golden$^{1,*}$, Daniel O'Malley$^{1,2}$\\
{\footnotesize$^1$Computational Earth Science, Los Alamos National Laboratory, Los Alamos, NM 87545}\\
{\footnotesize$^2$Department of Computer Science and Electrical Engineering, University of Maryland,}\\{\footnotesize Baltimore County, MD 21250}\\{\footnotesize$^*$ Corresponding author: golden@lanl.gov}}

\begin{document}

\maketitle

\begin{abstract}
It was recently shown that quantum annealing can be used as an effective, fast subroutine in certain types of matrix factorization algorithms.
The quantum annealing algorithm performed best for quick, approximate answers, but performance rapidly plateaued.
In this paper, we utilize reverse annealing instead of forward annealing in the quantum annealing subroutine for nonnegative/binary matrix factorization problems.
After an initial global search with forward annealing, reverse annealing performs a series of local searches that refine existing solutions.
The combination of forward and reverse annealing significantly improves performance compared to forward annealing alone for all but the shortest run times.   
\end{abstract}

\section{Introduction}

Due to the slowing progress of classical computing \cite{danowitz2012cpu}, new computational architectures \cite{geer2005chip,owens2008gpu,monroe2014neuromorphic} have gained much interest in recent years.
One such architecture is quantum annealing \cite{kadowaki1998quantum}.
Recently, D-Wave's quantum annealing hardware \cite{johnson2011quantum,gibney2017d} has introduced a new form of annealing -- reverse annealing \cite{chancellor2017modernizing,ohkuwa2018reverse}.
Here, we explore the use of reverse annealing in the context of Nonnegative/Binary Matrix Factoriztion (NBMF), which has shown some promise in combination with quantum annealing \cite{omalley2018nbmf}.

The NBMF algorithm factors a matrix $A$ into the product of a nonnegative, real-valued matrix $B$ and a binary matrix $C$.
The NBMF algorithm is a variant of the Nonnegative Matrix Factorization (NMF) algorithm (which allows $C$ to be real-valued rather than just binary). NMF, and by extension NBMF, are useful in machine learning contexts that seek to decompose a large data set into a set of features along with a mixing matrix, e.g. learning facial features \cite{lee1999learning}, text mining \cite{pauca2004mining}, and hyperspectral imaging \cite{rajabi2014spectral}. 
Our implementation of the NBMF algorithm employs an alternating least-squares approach, where each iteration includes the solution of a binary least squares problem and a nonnegative least squares problem.
The binary least squares problem is solved with the quantum annealer.
It has been shown that a quantum annealer provided noticeable speed-up compared to two classical solvers for the binary least squares problem \cite{omalley2018nbmf}.

One downside of the quantum annealer approach is that improvement in solution quality from iteration to iteration quickly plateaus.
This is because the forward annealing approach that was used previously could only perform global searches when solving the binary least squares problem.
This ignores the results of the solutions from previous iterations, which is likely a good starting point for the next iteration. 
Instead, the annealing process almost always produces a factor matrix that is very different from the factor matrix at the previous iteration.
In practice, this means that the algorithm hops around solution space at random, quickly finding good solutions but never refining them beyond a certain level of accuracy.

Fortunately, the latest iteration of the D-Wave hardware, the 2000Q, allows us to explore solutions around some initial classical state. 
This process is known as reverse annealing. 
In this paper, we utilize reverse annealing to improve performance of the NBMF algorithm. 
Specifically, we use reverse annealing to explore local minima near an initial state defined by the results of the previous iteration of the algorithm. 
This significantly reduces the iteration-over-iteration change in the algorithm, allowing promising solutions to be refined rather than discarded. 

\section*{Review of NBMF algorithm}\label{sec:review}
The NBMF algorithm takes a real-valued $n \times m$ matrix $A$ and finds $B$ and $C$ such that
\begin{equation}
A \approx B C
\end{equation}
where $B$ is a nonnegative $n \times k$ matrix and $C$ is a binary $k \times m$ matrix.
Generally, a small value of $k$ is used, so that the factorization is low rank.
The chief benefit of NBMF, as opposed to the more general nonnegative matrix factorization (NMF) algorithm (which only requires $C$ be nonnegative, rather than binary), is that $C$ tends to be sparse \cite{omalley2018nbmf}.
Additionally, the memory needed to store $C$ is low due to its binary nature. 

We will now give an outline of the NBMF algorithm and the implementation on the D-Wave; for full details of the algorithm see \cite{omalley2018nbmf}. 
After randomly initiating a seed matrix $C^{(0)}$, each iteration follows an alternating least squares approach:
\begin{align}
& \text{find } B^{(i)} = \text{arg min} \| A - X C^{(i-1)}\|, \\
& \text{find } C^{(i)} = \text{arg min} \| A - B^{(i)} X\| \label{eq:binary-matrix-opt}
\end{align}
An important feature of the NBMF algorithm is that Eq~\ref{eq:binary-matrix-opt} can be efficiently implemented on a quantum annealer. 
We used the D-Wave 2000Q quantum annealer at Los Alamos National Laboratory, which is designed to solve quadratic unconstrained binary optimization (QUBO) problems, generically formulated as
\begin{equation}\label{eq:nbmf-qubo}
	f(\vec{q}) = \sum a_i q_i + \sum b_{ij}q_i q_j.
\end{equation}
The D-Wave quantum annealer can be thought of as a sampler returning samples $\vec{q}$ drawn from a Boltzmann distribution where $f(\vec{q})$ is the energy. 
As we will discuss in the next section, the annealing process by which the D-Wave produces these samples is amenable to significant tuning by the user, with the goal of reducing the average energy of the samples. 

Returning to our matrix factorization problem, the columns $C_j$ in Eq~\ref{eq:binary-matrix-opt} can be solved for independently,
\begin{equation}
	C_j = \text{arg min} \| A_j - B \cdot \vec{q}\|.
\end{equation}
This can be converted into the QUBO format, Eq~\ref{eq:nbmf-qubo}, with the variable assignments 
\begin{eqnarray}
	a_j &=& \sum_l B_{lj}(B_{lj} - 2 A_{lj}), \\ 
	b_{jk} &=& 2 \sum_l B_{lj} B_{lk}.
\end{eqnarray}
In order to translate this QUBO onto the D-Wave hardware, we employ an embedding that chains multiple physical qubits into logical qubits.
This is required as the D-Wave hardware graph features limited connectivity, while the QUBO under study here requires a complete graph with $k$ vertices. 
The limited number of qubits and amount of connectivity on the physical hardware is in fact not a significant problem, as the goal of NBMF is to factor matrices in to a $C$ with small rank.

For consistency, we use the same data set (2,429 facial images) and rank $k=35$ as previous work \cite{omalley2018nbmf, lee1999learning}. 
The embedding, found via D-Wave's heuristic embedder, utilizes 437 physical qubits, with at most 15 physical qubits combined in to a single logical qubit. 
The facial images are composed of $19 \times 19 = 361$ greyscale pixels.
Our matrix $A$ is therefore of size $2429 \times 361$, with each column composed of the greyscale values of the pixels for an individual face. 

\section*{Reverse annealing: method, calibration \& timing}\label{sec:calibration}

The original implementation of NBMF on the D-Wave used the standard forward anneal procedure, where the device starts in an equal superposition of all possible states. 
Our motivation for this study is to use a new feature of the D-Wave 2000Q, reverse annealing, in order to improve performance. 
Reverse annealing begins in a specified classical state, then explores solutions in the local vicinity of that initial state. 
This allows us to iteratively improve upon solutions from previous iterations of the algorithm, rather than conducting global searches at every step. 

The NBMF algorithm begins with a random initialization of the $B$ and $C$ matrices. 
Reverse annealing from this random starting point is ineffective and requires many iterations to achieve results comparable to forward annealing. 
Conducting a single iteration with forward annealing and then switching to reverse annealing produces much better results (increasing the number of forward annealing iterations beyond this does not offer noticeable improvements).
This is in line with the idea that forward annealing performs a global search and reverse annealing performs a local search.
Intuitively, it is advantageous to start with a global search and then transition to a local search.
In the rest of this section we calibrate the reverse anneal process after the initial round of forward anneals. 

In the simplest terms, the reverse anneal process depends on two parameters:
\begin{itemize}
	\item reversal distance $r \in (0,1]$,
	\item reversal time $t_r$ (in $\mu$s).
\end{itemize}
The (dimensionless) reversal distance $r$ controls the scope of the local search. 
Setting $r=1$ is equivalent to conducting a global search (i.e. losing all information about the initial classical state), while setting $r=0$ will not conduct any search and instead return the initial state.
The reversal time $t_r$ controls how long the search is conducted. 
A longer search has a greater chance of returning a lower energy sample, but at the cost of slowing the algorithm down. 
For this analysis, we do not employ any other tuning of the D-Wave, e.g. spin reversal transforms. 

\begin{figure}[!ht]
\includegraphics[width = \linewidth]{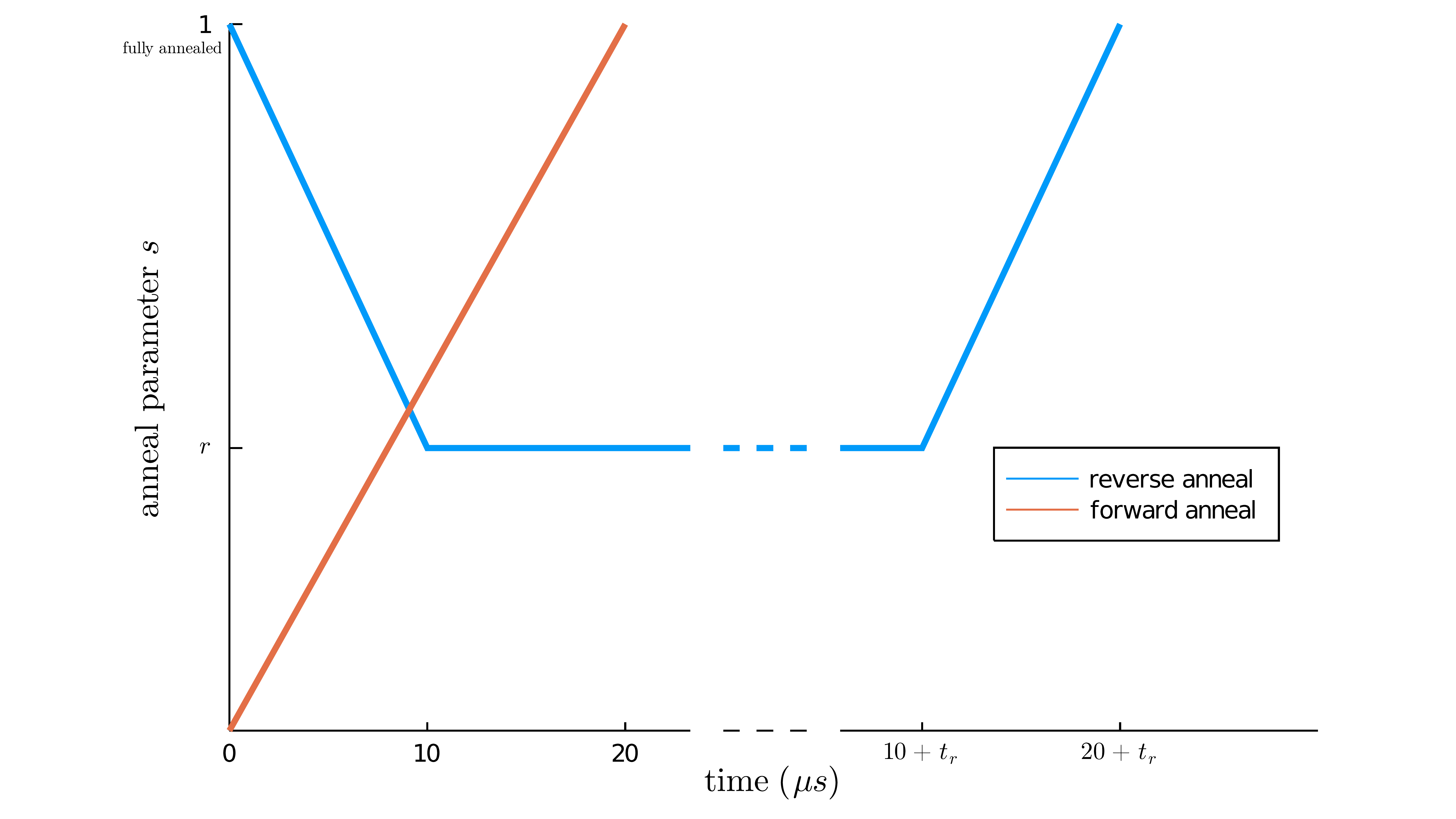}
\caption{{\bf Represention of the anneal schedules used as a function of the dimensionless anneal parameter $s$.}
 $s=1$ indicates a fully annealed system. The reverse anneal schedule (blue) is parameterized by the reversal distance $r$ and reversal time $t_r$, while default forward anneal schedule (orange) increases from $s=0$ to $s=1$ over 20 $\mu$s.}
\label{fig:anneal-schedule}
\end{figure}

For a given reversal distance $r$ and time $t_r$, the anneal schedule we use is 
\begin{equation}\label{eq:anneal-schedule}
	[(0,1),~(10,1-r),~(10+t_r, 1-r),~(20+t_r, 1)],
\end{equation}
where the first entry in each tuple is the elapsed time (in microseconds) and the second entry is the dimensionless anneal parameter $s \in [0,1]$, which controls the strength of a transverse magnetic field in the annealing device. 
As $s\to1$, the transverse field strength decreases, encouraging quantum tunneling towards the ground state of the QUBO.
The physical interpretation of Eq~\ref{eq:anneal-schedule} is that we begin in a specific annealed state at $t=0$, ``warm'' the system up to a certain temperature (parametrized by $r$), hold the system at that temperature for a time $t_r$, and then re-anneal the system.
See Fig.~\ref{fig:anneal-schedule} for a diagrammatic representation of the reverse anneal schedule, as parameterized by $r$ and $t_r$, and the D-Wave default forward anneal schedule.  

In addition to specifying the reverse anneal schedule, we must also specify the initial state. 
As discussed in the introduction, the NBMF algorithm naturally provides an initial configuration based on the results of the previous iteration of the algorithm. 
Specifically, if we are solving for $C^{(i)}_j$, i.e. the $i$th iteration of the $j$th column of $C$, we can use $C^{(i-1)}_j$ as the beginning point of our reverse anneal process. 

We characterize the efficacy of a reverse anneal sample by seeing if it:
\begin{itemize}
	\item is the same as the initial state,
	\item has a lower energy than the initial state (good),
	\item has a higher energy than the initial state (bad).
\end{itemize}
The frequency with which samples fall in to each category gives us an idea of how effective the reverse anneal is at finding improved solutions.

\begin{figure}[!ht]
\includegraphics[width = \linewidth]{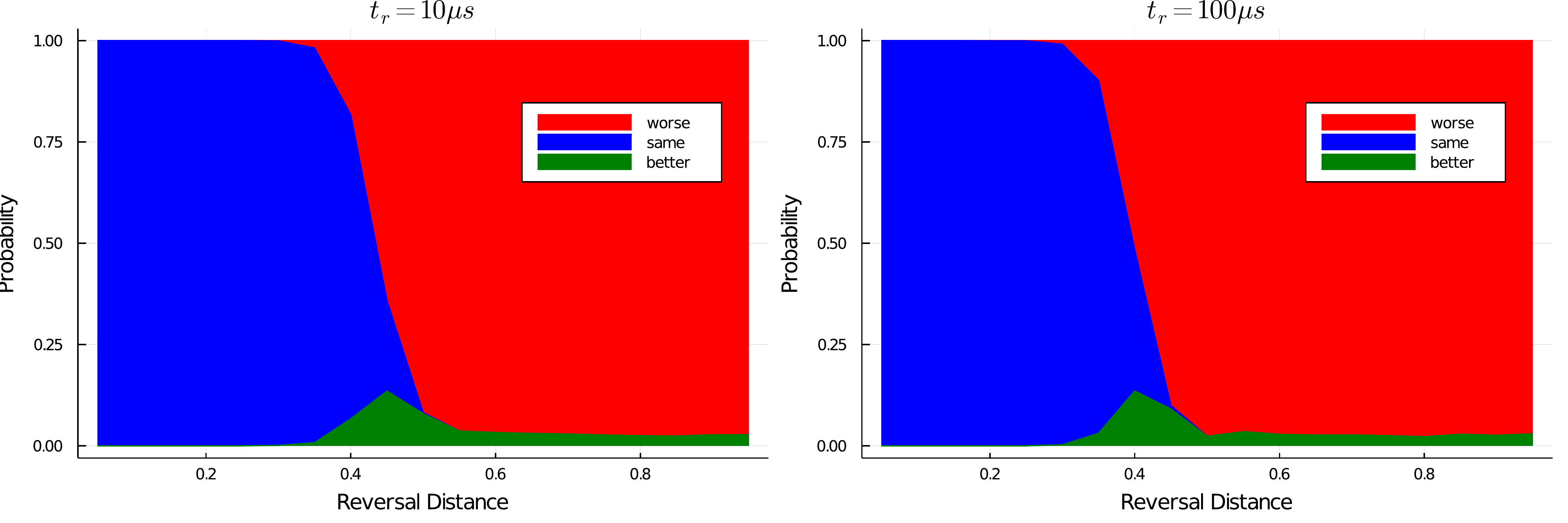}
\caption{{\bf Calibration of the reverse anneal process.} 
Evaluated for 100 randomly chosen QUBOs appearing during an evaluation of the NBMF algorithm for with $t_r = 10\mu s$ and $t_r = 100\mu s$. For a given reversal distance, the height of the green area indicates the mean probability that a reverse anneal sample will have a lower energy than the initial state. The heights of the red and blue areas indicate the mean probability that a given sample will be worse or the same, respectively, as the initial state. Standard deviations, not shown here, were upwards of $100\%$ of the mean, however $t_r = 10\mu s$ and $t_r = 100 \mu s$ always behaved similarly, and the peak reversal distance varied little from sample to sample.}
\label{fig:reverse-calibration} 
\end{figure}

We studied the effects of reversal time $t_r$ and reversal distance $r$ on 100 randomly selected QUBOs generated during an evaluation of the NBMF algorithm (after the first round of forward anneals). 
The overall effectiveness of reverse annealing on an individual QUBO was highly instance-specific; Fig~\ref{fig:reverse-calibration} shows the average of our results (the standard deviation for each point was generally near 100\% of the mean).
However, the impact of $t_r$ and $r$ remained consistent across instances. 
First, in increasing $t_r$ does not significantly increase the likelihood of discovering better states.
The peak probability of discovering a lower-energy sample was $13.5\% \pm 9.1\%$ for $t_r = 10\mu s$ and $13.52\% \pm 9.8\%$ for $t_r = 100 \mu s$.
Second, the peak reversal distance remained constant across samples ($r=0.45$ for $t_r = 10\mu s$ and $r=0.4$ for $t_r = 100\mu s$).
In order to minimize QPU access time, we therefore adopt a reverse anneal schedule for the NBMF algorithm with $r = 0.45$ and $t_r = 10\mu s$.

We note that the choice of these parameters has some dependence on the matrix that is being factored.
For example, the same calibration procedure evaluated on a matrix with random values (as opposed to the highly structured facial imagery data) revealed an optimal reversal distance of $r=0.2$. 

There is additional computational overhead related to the reverse anneal process, such as configuring the hardware in to the chosen initial state before each anneal.
Therefore, for the purposes of comparing forward and reverse anneal efficacy we will look at quality of solution vs. total QPU access time (as opposed to (annealing time $\times$ number of anneals), as was done in \cite{omalley2018nbmf}). 
The total QPU access time is calculated via
\begin{multline}
	\text{total QPU access time = (anneal + readout + delay)}\times\text{number of samples} \\ \text{+ QPU programming time} 
\end{multline}
Forward and reverse anneals share identical readout and QPU programming times ($123\mu$s and $8001\mu$s, respectively).
As previously discussed, the forward anneal takes $20\mu$s while the reverse anneal takes $30\mu$s. 
The major difference is in the `delay' time, as this is the period when the quantum annealer is reset to the initial state between anneals. 
For the D-Wave 2000Q used for this study, located at Los Alamos National Laboratory, the delay time per sample in the forward anneal case is $21\mu$s, while the delay time per reverse anneal sample is $520\mu$s. 
So we see that the biggest time commitment in doing reverse anneals comes not from the longer anneal schedule but from the repeated state preparation. 

When comparing the reverse anneal results against the original forward anneal version of the algorithm, we allot each method equal QPU access time. 
Given the timing values discussed above, we find that the ratio
\begin{equation}\label{eq:qpu-timing}
	\text{number of reverse anneals} = 0.24*(\text{number of forward anneals})
\end{equation}
results in equivalent total QPU access time.
The remaining important variables are the number of anneals per QUBO and the total number of iterations for the algorithm to run. 
We discuss these in the following section.

\section*{Results}\label{sec:results}
In this section we use reverse annealing in the NBMF algorithm to factor the dataset of 2,429 facial images studied in \cite{omalley2018nbmf} in to a $2429 \times 35$ non-negative matrix $B$ and $35 \times 2429$ binary matrix $C$. 
In this application, the columns of the $C$ matrix can be interpreted as decompositions of each face in to 35 component features. 
First, we will examine the differences between the two algorithms for a fixed number of anneals.
We will then study the efficacy of the two algorithms as a function of total QPU access time.

\begin{figure}[!ht]
\includegraphics[width = \linewidth]{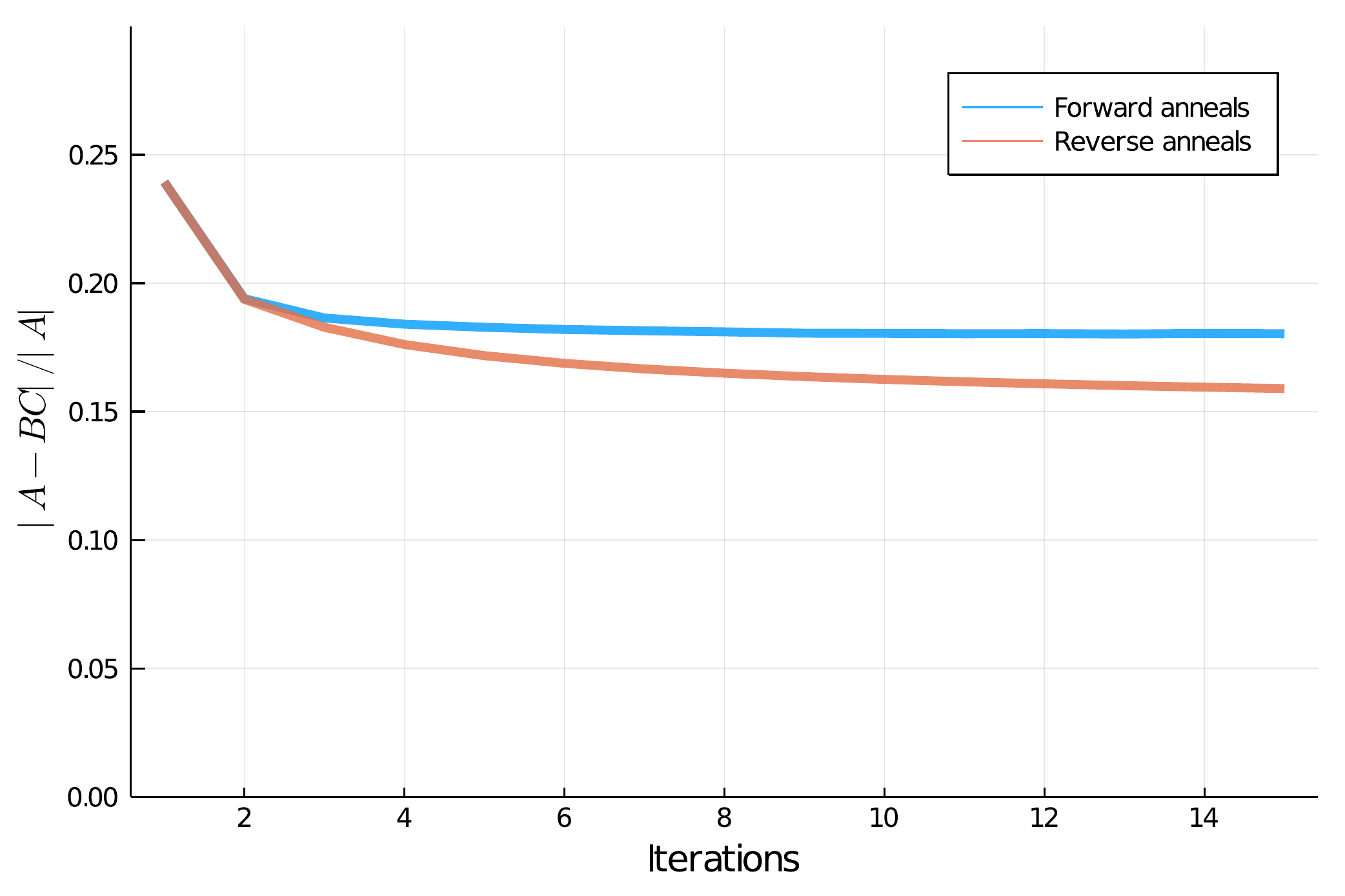}
\caption{{\bf Comparison of forward and reverse anneal versions of the NBMF algorithm for fixed annealing time.} 
Mean performance reported from five evaluations of the forward and reverse annealing versions of the NBMF algorithm, with 1000 forward anneals and 240 reverse anneals per QUBO, corresponding to a total QPU access time of 6182 seconds over the full evaluation of each algorithm. Standard deviation (not shown) was less than 1\% of the mean.}
\label{fig:forward-vs-reverse-abc}
\end{figure}

Fig~\ref{fig:forward-vs-reverse-abc} shows the results of the two algorithms with 6182 seconds of total QPU access time (equivalent to 1000 forward anneals or 240 reverse anneals per QUBO).
The reverse anneal algorithm shows consistent improvement for many more iterations, and produces a better result than forward anneal by the third iteration.

\begin{figure}[!ht]
\includegraphics[width = \linewidth]{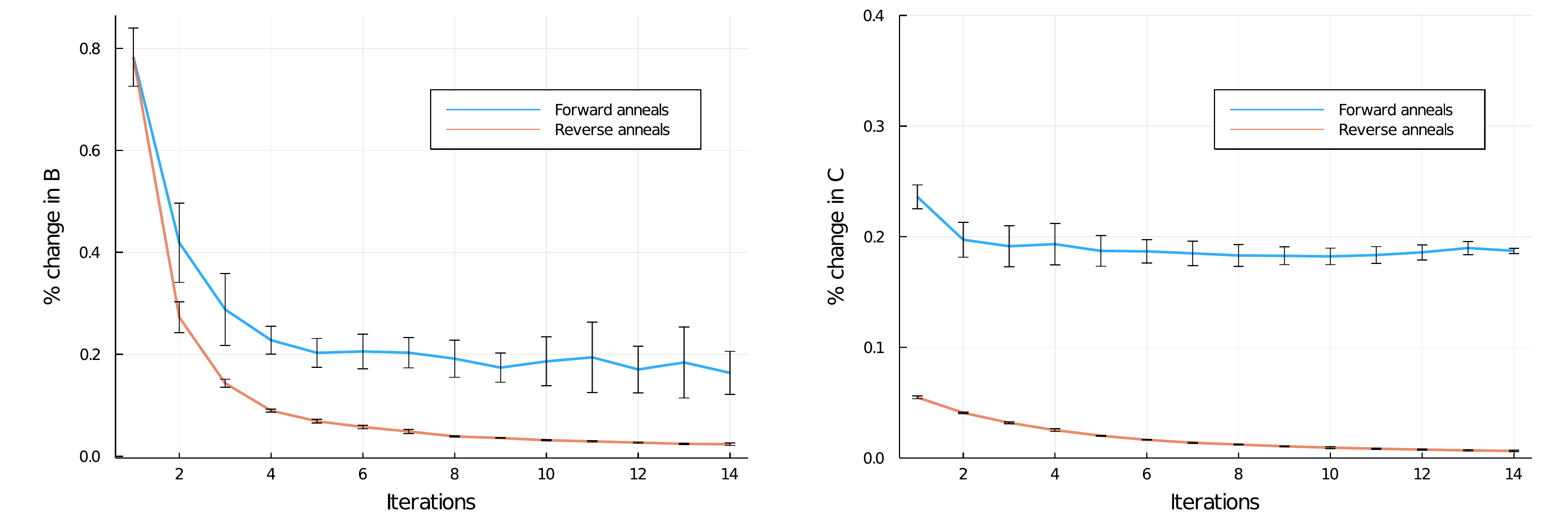}
\caption{{\bf Iterative improvement of $B$ and $C$ matrices for forward and reverse anneal versions of the NBMF algorithm.} 
Data taken during the evaluation of the algorithms as described in Fig~\ref{fig:forward-vs-reverse-abc}.}
\label{fig:forward-vs-reverse-bc}
\end{figure}

Recall that our hypothesis, outlined in the introduction, is that reverse annealing will outperform forward annealing due to more refinement of existing solutions as opposed to generation of entirely new solutions per QUBO. 
If we define
\begin{equation}
	\text{\% change in } B = \frac{\| B^{(i+1)} - B^{(i)}\|}{\|B^{(i)}\|}
\end{equation}
and \% change in C as the Hamming distance between $C^{(i+1)}$ and $C^{(i)}$ divided by the size of $C$, then we can look at the iteration-over-iteration change in the $B$ and $C$ matrices to see if this is indeed the case, see Fig~\ref{fig:forward-vs-reverse-bc}.
The iterative improvement in $C$ is particularly striking and shows that the forward anneal is constantly changing the $C$ matrix while reverse anneal pushes towards a local minimum.

\begin{figure}[!ht]
\includegraphics[width = \linewidth]{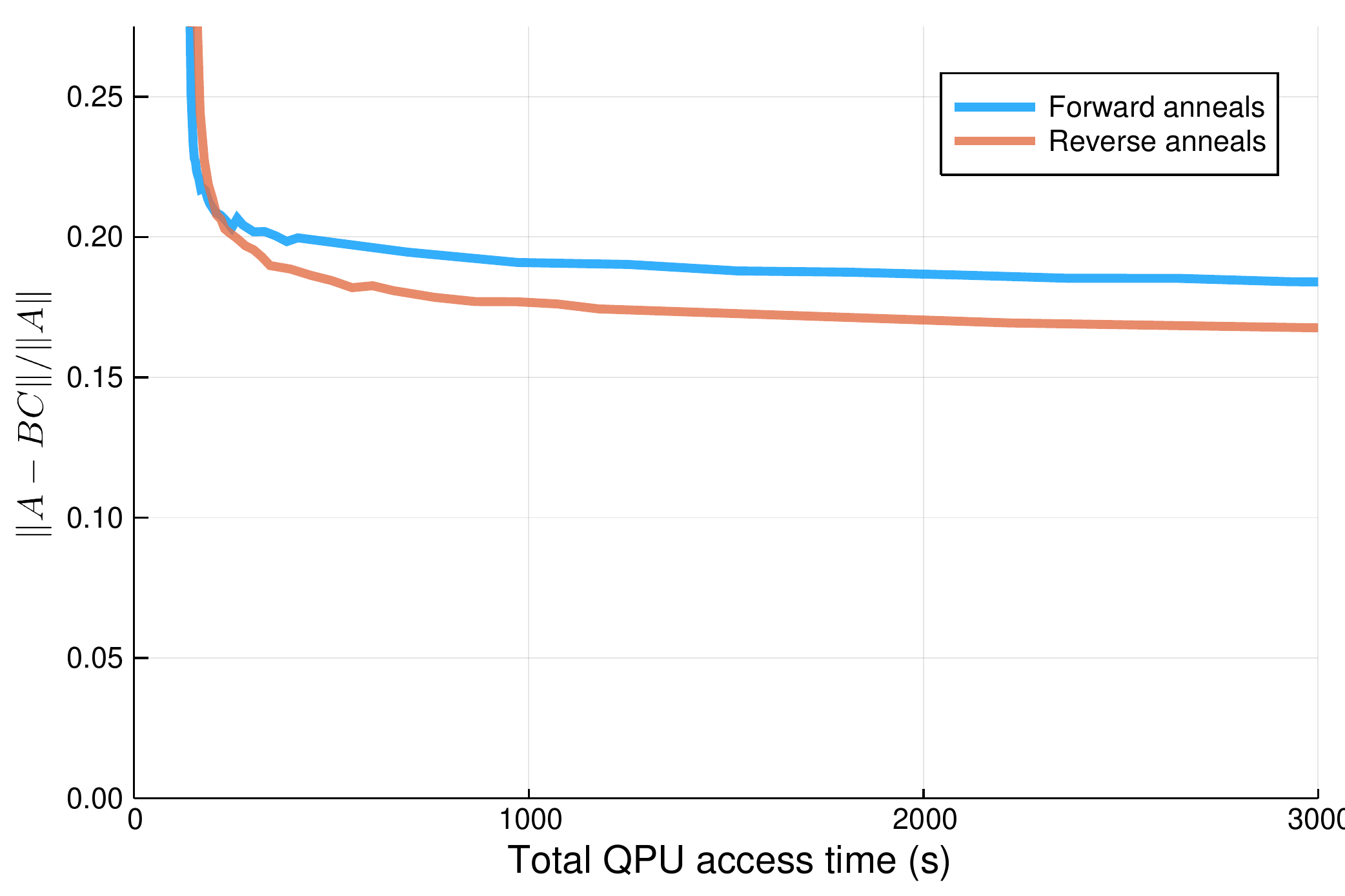}
\caption{{\bf Comparison of forward and reverse anneal versions of NBMF algorithm after seven iterations for a range of annealing times.}
Data taken from 725 distinct evaluations of the NBMF algorithm, varying the number of forward and reverse anneals per evaluation. Reverse annealing results in up to a $12\%$ increase in performance.}
\label{fig:forward-vs-reverse-abc2}
\end{figure}

In Fig~\ref{fig:forward-vs-reverse-abc2} we compare the efficacy of the two algorithms over multiple values of QPU access time.
Here we see that for very small values of total QPU access time, forward annealing results in superior performance.
However, once the total QPU access time exceeds $\approx 210$s, which corresponds to 7 reverse anneals per QUBO, reverse annealing overtakes forward annealing, eventually plateauing at approximately 12\% improvement over the forward annealing algorithm.  

The reason that forward annealing outperforms reverse annealing for small sample size is straightforward. 
For a single reverse anneal sample, the likelihood of finding a better state than the initial configuration is quite low ($\le 25\%$, sometimes much lower, see Fig~\ref{fig:reverse-calibration}). 
Therefore, when the number of reverse anneals per QUBO is small, most iterations will result in no change.
Forward annealing might be finding worse or different solutions for each QUBO, but the fact that they are new solutions for each iteration means that overall the algorithm can improve. 
As the number of reverse annealing samples increases, the chance of finding a better/different solution increases significantly, resulting in improved performance. 

\begin{figure}[!ht]
\includegraphics[width = \linewidth]{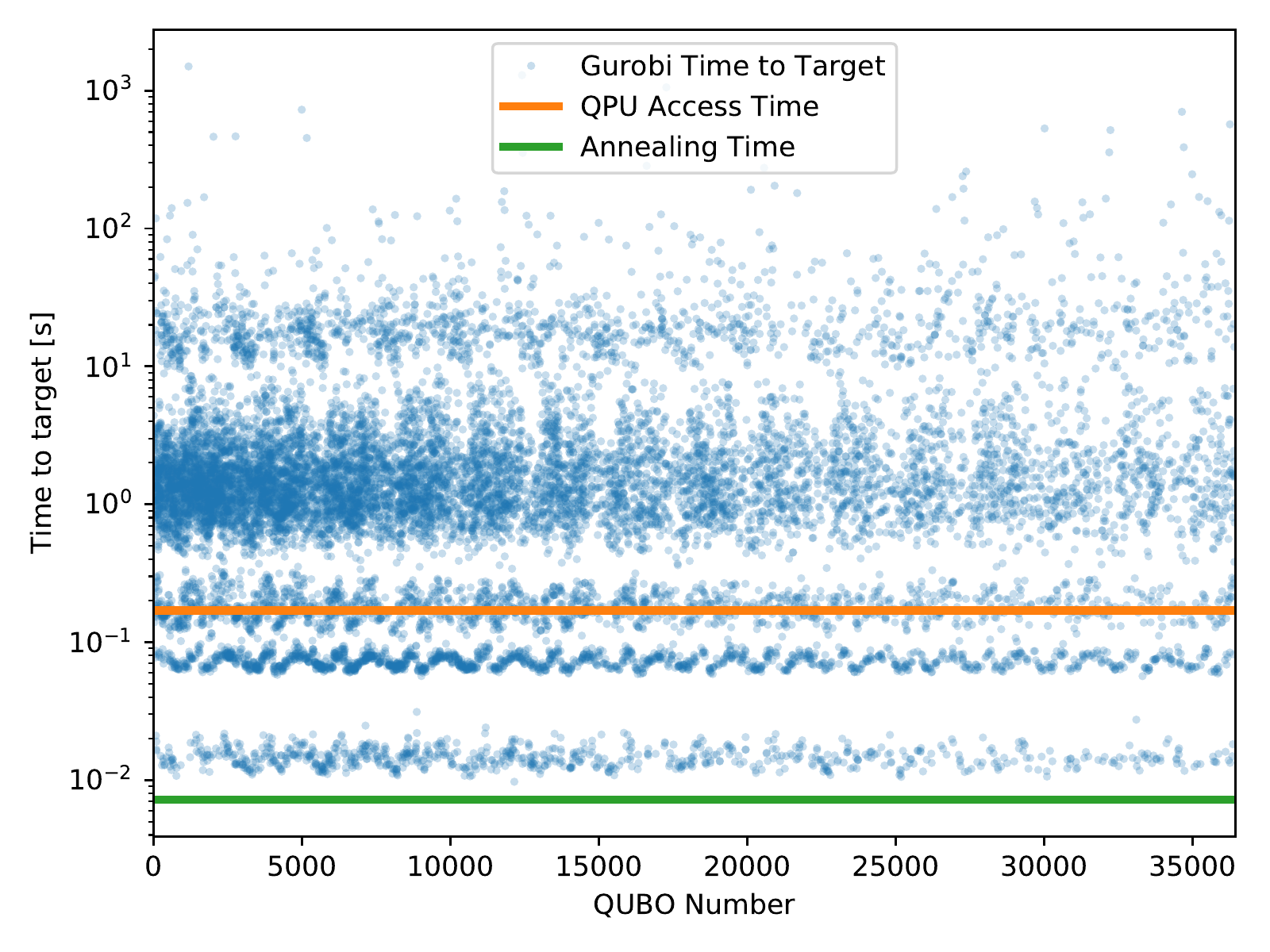}
\caption{{\bf Comparison between Gurobi and reverse annealing.}
The time required for Gurobi to find a solution that is as good as or better than the solution found by reverse annealing for each QUBO in a factorization problem is shown in comparison to the QPU access time and the annealing time.
Cases where reverse annealing failed to find a better solution are excluded from the plot.
In many cases, the time required by Gurobi exceeds both the annealing time and the QPU access time.}
\label{fig:gurobi-benchmark}
\end{figure}

We performed a benchmark using Gurobi that is similar to the benchmark performed previously for this problem \cite{omalley2018nbmf} (Fig~\ref{fig:gurobi-benchmark}).
The basic mechanism of the benchmark is to determine the time Gurobi requires to find a solution that is as good as or better than the solution found by reverse annealing.
We call this time the ``time to target.''
For this test we used data from one of the runs described in Fig~\ref{fig:forward-vs-reverse-abc}, i.e. 240 reverse anneals per QUBO.
The most notable change from the previous incarnation of this benchmark is that Gurobi uses the classical state that reverse annealing starts in as a starting point to perform its optimization, whereas it was not given a starting point in the previous benchmark.
This was done to put it on more equal ground with the reverse annealing algorithm.
The other changes are due to different classical hardware being used (a 2.4GHz 8-core Intel Core i9 processor) and an updated version of Gurobi (version 9.0.2).
We ran Gurobi using 8 threads, whereas previously 1 thread was used.
Fig.~\ref{fig:gurobi-benchmark} shows the result of the benchmark on each of the QUBOs where reverse annealing was performed in the course of a matrix factorization.
Overall, the benchmark indicates that reverse annealing is performing well.
Note that cases where reverse annealing did not improve on the previous solution were assigned a time to target of 0, and are not included in the plot.
The sum of the individual time to targets was $\sim$68000 seconds.
For reference, the total annealing time was 262 seconds and the total QPU access time was 6182 seconds.

\section*{Conclusion}\label{sec:conclusion}
The results of this work suggest that reverse annealing improves the quality of the NBMF factorization by 12\% for this application. 
This improvement is seen when the number of reverse anneals evaluated per QUBO is at least 7 (which is equivalent in QPU access time to 29 forward anneals).
In \cite{omalley2018nbmf}, it was observed that quantum annealing had the largest performance gains relative to classical benchmarks in the short annealing timeframe, $\mathcal{O}(10)$ forward anneals per QUBO.
Reverse annealing improves performance in the longer annealing timeframe, thus further establishing quantum annealing as a strong approach for non-negative binary matrix factorization.

In addition to characterizing the performance in terms of the quality of the factorization given a fixed time, it could be characterized in terms of how long it takes to obtain a factorization of a given quality.
By this standard, reverse annealing would also perform well once the quality of the factorization is set sufficient low.
Since NBFM with forward annealing tends to plateau at a worse factorization quality, the speed-up with reverse annealing would be very large once the factorization quality is set beyond this plateau.

Our results could be improved upon in several ways. 
First, it is possible that the optimal reverse anneal schedule could depend on how many iterations have already occured (i.e., as better solutions become harder to find). 
It is also our hope that future quantum annealing hardware will feature more rapid state initialization, as this accounts for over $98\%$ of the additional time related to reverse annealing. 
This would improve the performance of NBMF with reverse annealing but leave the performance of NBMF with forward annealing unchanged.
Lastly, the exact nature of the matrix being factorized appears to play a role in determining how effective the algorithm is, and this could be explored further.


\begin{thebibliography}{10}

\bibitem{danowitz2012cpu}
Andrew Danowitz, Kyle Kelley, James Mao, John~P Stevenson, and Mark Horowitz.
\newblock Cpu db: recording microprocessor history.
\newblock {\em Communications of the ACM}, 55(4):55--63, 2012.

\bibitem{geer2005chip}
David Geer.
\newblock Chip makers turn to multicore processors.
\newblock {\em Computer}, 38(5):11--13, 2005.

\bibitem{owens2008gpu}
John~D Owens, Mike Houston, David Luebke, Simon Green, John~E Stone, and
  James~C Phillips.
\newblock {GPU} computing.
\newblock {\em Proceedings of the IEEE}, 96(5):879--899, 2008.

\bibitem{monroe2014neuromorphic}
Don Monroe.
\newblock Neuromorphic computing gets ready for the (really) big time.
\newblock {\em Communications of the ACM}, 57(6):13--15, 2014.

\bibitem{kadowaki1998quantum}
Tadashi Kadowaki and Hidetoshi Nishimori.
\newblock Quantum annealing in the transverse ising model.
\newblock {\em Physical Review E}, 58(5):5355, 1998.

\bibitem{johnson2011quantum}
Mark~W Johnson, Mohammad~HS Amin, Suzanne Gildert, Trevor Lanting, Firas Hamze,
  Neil Dickson, R~Harris, Andrew~J Berkley, Jan Johansson, Paul Bunyk, et~al.
\newblock Quantum annealing with manufactured spins.
\newblock {\em Nature}, 473(7346):194--198, 2011.

\bibitem{gibney2017d}
Elizabeth Gibney.
\newblock D-wave upgrade: How scientists are using the world's most
  controversial quantum computer.
\newblock {\em Nature}, 541(7638):447--448, 2017.

\bibitem{chancellor2017modernizing}
Nicholas Chancellor.
\newblock Modernizing quantum annealing using local searches.
\newblock {\em New Journal of Physics}, 19(2):023024, 2017.

\bibitem{ohkuwa2018reverse}
Masaki Ohkuwa, Hidetoshi Nishimori, and Daniel~A Lidar.
\newblock Reverse annealing for the fully connected p-spin model.
\newblock {\em Physical Review A}, 98(2):022314, 2018.

\bibitem{omalley2018nbmf}
Daniel O’Malley, Velimir~V Vesselinov, Boian~S Alexandrov, and Ludmil~B
  Alexandrov.
\newblock Nonnegative/binary matrix factorization with a d-wave quantum
  annealer.
\newblock {\em PloS one}, 13(12):e0206653, 2018.

\bibitem{lee1999learning}
\newblock Daniel~D Lee and H Sebastian Seung.
\newblock Learning the parts of objects by non-negative matrix factorization.
\newblock {\em Nature}, 401(6755):788--791, 1999.

\bibitem{pauca2004mining}
\newblock V.Paul Pauca, Farial Shahnaz, Michael Berry, and Robert Plemmons. 
\newblock Text Mining Using Non-Negative Matrix Factorizations.
\newblock SIAM Proceedings Series. 10.1137/1.9781611972740.45, 2004. 

\bibitem{rajabi2014spectral}
\newblock Roozbeh Rajabi and Hassan Ghassemian.
\newblock Spectral Unmixing of Hyperspectral Imagery Using Multilayer NMF.
\newblock {\em IEEE Geoscience and Remote Sensing Letters}, 12(1):38--42, 2015.

\end{thebibliography}
\end{document}